\documentclass[letter, 10 pt, conference]{ieeeconf}  
\IEEEoverridecommandlockouts                              

\usepackage{hyperref}
\usepackage{placeins}
\usepackage{amssymb}
\usepackage{pifont}
\usepackage{amsmath}
\usepackage{bm}
\usepackage{cancel}
\usepackage{units}
\usepackage{multirow}
\usepackage{ulem}

\usepackage{tikz}
\usetikzlibrary{external,arrows,shapes,automata,backgrounds,petri,fit,arrows.meta,positioning,calc,matrix,chains,scopes}
\usepackage[skins]{tcolorbox}
\usepackage{pgfplotstable}
\usepackage{pgfplots}
\usepackage{subcaption}

\newcommand{\timeCricitical}{t_\text{0}}
\newcommand{\timeBefore} {t_\text{lb}}
\newcommand{\Occupancy}{\bm{G}_{t}}

\newcommand{\Occupancyfirst}{\bm{G}_{t_{\text{lb}}+\Delta t}}
\newcommand{\Occupancylast}{\bm{G}_{t_\text{0}}}

\title{\LARGE \bf
Open-set Recognition based on the Combination of Deep Learning and Ensemble Method for Detecting Unknown Traffic Scenarios
}
\newcommand\copyrighttext{\footnotesize \textcopyright 2021 IEEE. Personal use of this material is permitted. Permission from IEEE must be obtained for all other uses, in any current or future media, including reprinting/republishing this material for advertising or promotional purposes, creating new collective works, for resale or redistribution to servers or lists, or reuse of any copyrighted component of this work in other works.
DOI:\href{tba}{tba}}
\newcommand\copyrightnotice{%
\begin{tikzpicture}[remember picture,overlay]
\node[anchor=south,yshift=10pt] at (current page.south) {\fbox{\parbox{\dimexpr\textwidth-\fboxsep-\fboxrule\relax}{\copyrighttext}}};
\end{tikzpicture}%
} 

\author{Lakshman Balasubramanian$^{1}$, Friedrich Kruber$^{1}$, Michael Botsch$^{1}$ and Ke Deng$^{2}$
\thanks{$^{1}$Technische Hochschule Ingolstadt, Research Center CARISSMA, Esplanade
	10, 85049 Ingolstadt, Germany, {\tt\small firstname.lastname@thi.de}}%
\thanks{$^{2}$Royal Melbourne Institute of Technology, Melbourne, Australia, 
        {\tt\small firstname.lastname@rmit.edu.au}}%
}

\begin{document}

\maketitle
\copyrightnotice
\thispagestyle{empty}
\pagestyle{empty}

\begin{abstract}
An understanding and classification of driving scenarios are important for testing and development of autonomous driving functionalities. Machine learning models are useful for scenario classification but most of them assume that data received during the testing are from one of the classes used in the training. This assumption is not true always because of the open environment where vehicles operate. This is addressed by a new machine learning paradigm called open-set recognition. Open-set recognition is the problem of assigning test samples to one of the classes used in training or to an unknown class. This work proposes a combination of Convolutional Neural Networks (CNN) and  Random Forest (RF) for open set recognition of traffic scenarios. CNNs are used for the feature generation and the RF algorithm along with extreme value theory for the detection of known and unknown classes. The proposed solution is featured by exploring the vote patterns of trees in RF instead of just majority voting. By inheriting the ensemble nature of RF, the vote pattern of all trees combined with extreme value theory is shown to be well suited for detecting unknown classes. The proposed method has been tested on the highD and OpenTraffic datasets and has demonstrated superior performance in various aspects compared to existing solutions. 

\end{abstract}

\section{Introduction}

The advances in sensor technology and computing capabilities have made it possible to move towards higher degrees of vehicle autonomy~\cite{Winner2015}. SAE J3016~\cite{International2014} provides a step by step approach for vehicle autonomy by dividing it into five levels. With higher levels of automation, new and improved autonomous driving functions are added. Also, the system must be capable of handling different driving scenarios for those functions without the need of giving the control back to the driver. One of the main challenges for such a highly autonomous vehicle is to understand an encountered traffic scenario, as it is essential for testing and developing autonomous driving functionalities like path planning~\cite{Chaulwar2017} and behaviour planning. Statistical learning methods are shown to be useful for scenario categorization in~\cite{Cara2015,Beglerovic2019}. This is mainly because the transition of environment states in a traffic scenario is not completely random, there exist driving patterns for different types of scenarios because of infrastructure, traffic rules, etc.~\cite{Gindele2013}. Machine learning methods can learn the transitions  based on the data.

Most of the existing works assume that the data received by machine learning models during testing or after deployment is from one of the classes ($known$) in the training data. This is termed as \textit{closed-world assumption} in~\cite{Bendale2014}. However, in the real world, it is possible to have many new or $unknown$ scenario categories due to the open environment in which  vehicles operate. This leads to a new machine learning paradigm called Open-Set Recognition (OSR). OSR is the problem of identifying classes which are not explicitly labelled in the training dataset, i.\,e., identifying unknown classes. This research problem is important for autonomous driving applications, but highly challenging. 

According to \cite{Jain2014,Oza2019}, the performance of the classifiers, if trained under the closed-world assumption, will noticeably degrade if the assumption does not hold, no matter what models they are like Support Vector Machines (SVM) and CNN. Even though this is a critical task, unfortunately none of the publications known to the authors can classify traffic scenarios to one of the classes used in the training or as an unknown class. 

While there is a significant body of work on incremental learning algorithms that handle new instances of known classes~\cite{Rebuffi2016,Wu2019LargeSI}, OSR is less investigated. Current studies migrate the classifiers originally developed under closed-world assumption to solve OSR by introducing particular extensions. These classifiers include SVM~\cite{Scheirer2014},  and deep neural networks~\cite{Bendale2015,Shu2017,Oza2019a}. In these works, the main task is to detect data samples which are not from any of the known classes. Various methods have been proposed for this goal including distance metrics to known clusters \cite{Bendale2014,Mendes-Junior2016}, the reconstruction error \cite{Zhang2017,Oza2019a}, the Extreme Value Theory~(EVT)~\cite{Scheirer2014,Bendale2015}, and the synthesis of outliers from a generative model \cite{Ge2017,Neal2018}. 

As opposed to the above mentioned works which deal with image-based datasets, traffic scenarios are described as time series data. Hence, to perform open-set recognition of traffic scenarios and to improve the performance of the OSR task, this work proposes a combination of CNN and RF algorithm. The CNN acts as feature extractors. The RF algorithm along with EVT assigns a test sample to one of the classes used in the training or as an unknown class. The proposed solution is featured by exploring the vote patterns of trees in RF instead of majority voting. The vote patterns modelled using extreme value distributions is used to detect the unknown class. By inheriting the ensemble nature of RF and the vote patterns of all trees combined with extreme value theory is shown to be well suited for detecting unknown classes. The implementation of the architecture is made publicly available\footnote{https://github.com/lab176344/Traffic\_Sceanrios-VoteBasedEVT}. The contributions of this work are the following:

\begin{itemize}
    \item To our knowledge this work proposes for the first time the OSR task for traffic scenarios.
	\item  An ensemble learning method based on RF and a feature generation step implemented by 3D CNN is proposed for solving the OSR task.     
	\item Instead of taking the majority vote across trees, the vote patterns of trees in a RF are explored. The vote-based unknown class detection and the vote-based EVT model are used to detect the traffic scenarios not belonging to known classes.
	\item A comprehensive analysis is performed using the highD~\cite{Krajewski2018a} and the OpenTraffic~\cite{Kruber2020} datasets to test the robustness of the proposed method for the open-set recognition of traffic scenarios. The results are compared with other state-of-the-art methods and also ablation studies are performed. 
\end{itemize}

The remainder of the paper is organised as follows. Section~\ref{Sec:RelWorks} presents the related work. Section~\ref{Sec:Proposed} describes the proposed methodology and the rationale behind it, and Section~\ref{Sec:Exper} illustrates and discusses the experiment results. Section~\ref{Sec:Abla} presents the ablation studies. Finally, this paper is concluded in Section~\ref{sec:conclusion}. 

\section{Related Works}
\label{Sec:RelWorks}
\subsubsection{Traffic Scenario Classification}
Machine learning methods have been proposed for the task of scenario classification in previous studies. In~\cite{Reichel2010}, the authors propose a scenario based RF algorithm for classifying convoy merging situations. In~\cite{Beglerovic2019}, a CNN based method is used for classifying scenarios like a lane change, ego cut in-out, ego cut-out, etc. In~\cite{Khosroshahi2016}, a LSTM model is proposed for classifying scenarios in a four way intersection. Although all the above methods perform well for the intended task of scenario classification, they work under the closed-world assumption. Different from the above mentioned methods, this work proposes a solution for classifying a scenario to one of the classes used in the training or as an unknown class.

\subsubsection{Open-Set Recognition}
Current studies in OSR typically apply classifiers, which were originally developed under the closed-world assumption, by introducing particular extensions to detect data samples that deviate significantly from existing classes. These classifiers include SVM, nearest neighbour classifiers, and deep neural networks. In \cite{Scheirer2014}, a Weibull-calibrated SVM has been proposed, where the score calibration of SVM is performed using EVT. In \cite{Bendale2014}, an OSR method called \textit{nearest non-outlier} has been proposed based on the nearest mean classifier in which data samples undergo a Manhanalobis transformation and then they are associated with a class/cluster mean. In~\cite{Zhang2017}, an OSR method has been proposed based on a sparse representation which uses the reconstruction error as the rejection score for known classes.  

The OSR methods based on deep neural network can be divided into discriminative and generative model-based methods. The approaches in \cite{Bendale2015,Shu2017,Oza2019a} are examples of discriminative model-based methods. In~\cite{Bendale2015}, the softmax scores are calibrated using class specific EVT distributions. The re-calibrated scores are termed as \textit{Openmax} scores. The EVT distributions are used to model the euclidean distance between the activation vectors, i.\,e., the features from the penultimate layer of the network and the saved mean activation vectors for each class. 

In \cite{Shu2017}, the authors introduce a novel sigmoid-based loss function to train a deep neural network. The score generated using the sigmoid function is used as a threshold for OSR. In~\cite{Oza2019a}, an OSR method based on autoencoders has been proposed where the reconstruction error is used. Recently, generative model-based approaches have been investigated \cite{Ge2017,Neal2018}. In general, generative model-based approaches train a closed-set recognition model on $K+1$ classes, where $K$ classes are known from the original training data set and the unknown $(K+1)$-th class is synthesized by a generative model.

All the above mentioned methods are proposed for image-based datasets and here in this work traffic scenarios are time series data. Also, different from all existing methods, the proposed OSR solution in this work is an ensemble learning method based on RF trees which uses CNN for feature extraction. The vote patterns of trees in the RF are explored in the attempt to maintain more information for OSR. Each tree in the RF learns some unique insights about the data because of bootstrapping and random feature selection used to train the RF. So the vote patterns of trees together with extreme value distributions are robust and well-suited for OSR.  

\begin{figure*}
\vspace{2mm} 
	\centering
	\includegraphics[width=0.7\linewidth]{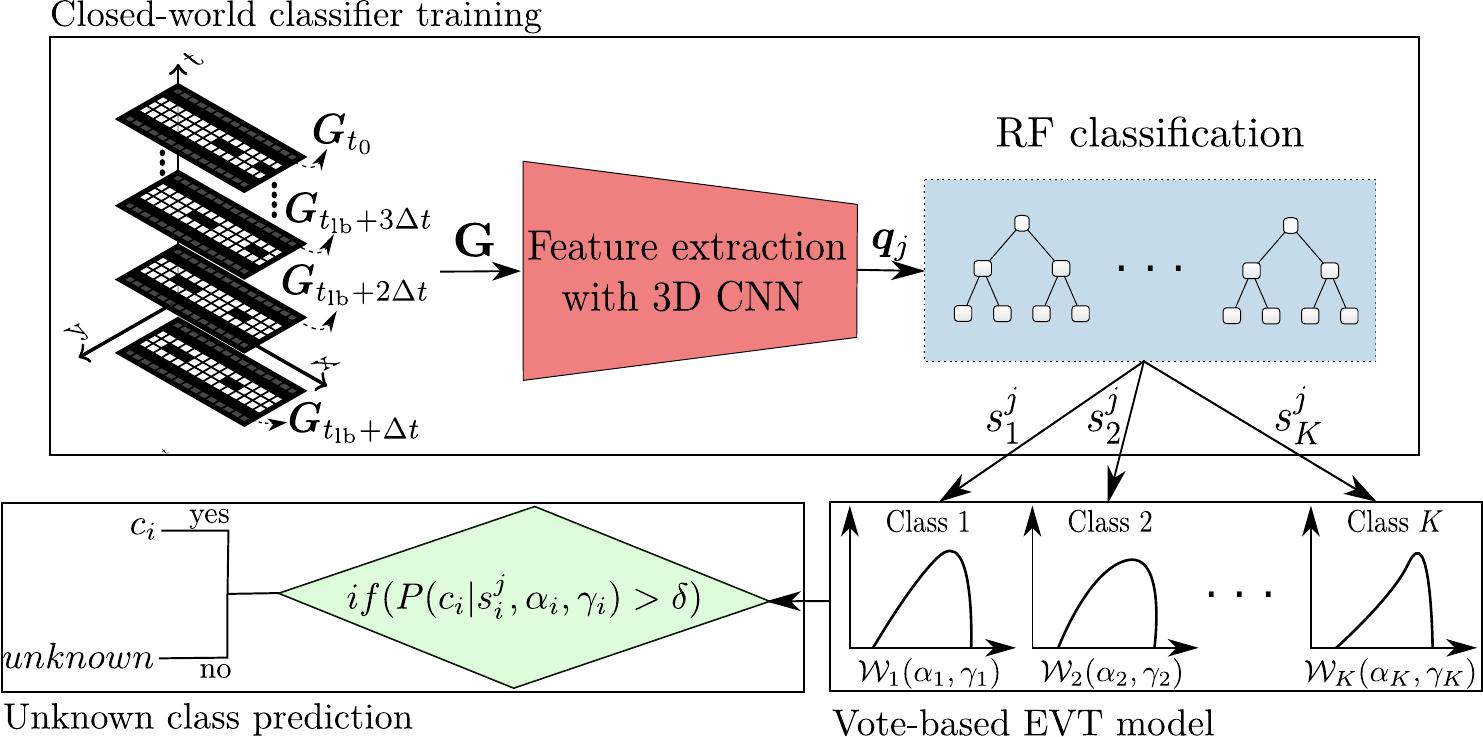}
	\caption{Open-set recognition architecture.}
	\label{Fig:Arch}
\end{figure*}

\section{Methodology}
\label{Sec:Proposed}
This section describes the proposed OSR architecture and each component is discussed in detail. In Fig.~\ref{Fig:Arch}, an overview of the architecture for the proposed OSR method is shown. Accordingly, the dataset is split into training dataset, calibration dataset and test dataset. The training and calibration data include samples of known classes only. The test data includes samples of both known and unknown classes. 

\subsection{Traffic Scenario Description}
\label{subsec:traffic}
The data analysed in this work is restricted to traffic scenarios where there the ego vehicle has a leading vehicle. The trigger used here to determine  the time  $\timeCricitical$  at which the traffic situation becomes interesting is determined based on the Time-Headway~(THW). 

The environment around the ego  should be represented in a compact and general way as possible. There are several options to define a traffic scenario. In~\cite{Beglerovic2019}, a polar occupancy map is proposed for the scenario description. In~\cite{Chaulwar2017}, the traffic scenario is represented as a sequence of occupancy grids. In~\cite{Ulbrich2015}, a context modelling approach is proposed for defining traffic scenarios.

A traffic scenario in this work is described similar to~\cite{Chaulwar2017} as a discretized space-time representation of the environment around the ego,  from the time $\timeBefore+\Delta t$ to  $\timeCricitical$. The $\timeBefore$ refers to a time before $\timeCricitical$, i.\,e., a time before the traffic situation becomes interesting. The time resolution is denoted with $\Delta t$. The traffic scenario at each time instance is represented as a 2D occupancy grid\footnote{A grid based representation of the environment with grid cells of pre-defined size filled with occupancy values}  $\Occupancy\in\mathbb{R}^{I \times J}$ with $I$ rows and $J$ columns. The occupancy probability of each cell $\Occupancy(i,j)$ is assigned either as 1 for occupied space, 0 for free space or 0.5 for an unknown region.  As shown in the Fig.~\ref{Fig:Arch}, a traffic scenario is represented as $\textbf{G}=\left[\Occupancyfirst,\ldots,\Occupancylast \right]$, where $\textbf{G}\in\mathbb{R}^{I\times J \times N_{\text{ts}}}$. The depth is $N_{\text{ts}}=1+\frac{t_\text{0}-t_\text{lb}}{\Delta t}$.
\subsection{Closed-World Classifier Training}\label{sec:closeset}
Let $\mathcal{D}=\{(\textbf{G}_1,y_1),(\textbf{G}_2,y_2),\ldots,(\textbf{G}_M,y_M)\}$ be the training dataset where $\textbf{G}$ is the scenario representation and $y\in\{c_1,c_2,\ldots,c_K\}$ is the corresponding label, one of $K$ classes. The task of closed-world training is to learn a prediction model which can accurately map $\textbf{G}$ to $\textsl{\textbf{{y}}}$, where $\textsl{\textbf{{y}}}\in\mathbb{R}^K$ is the one-hot representation of the predicted class. For example, the one-hot representation of class $c_2$ is $[0,1,0,\cdots,0]$. As shown in Fig. \ref{Fig:Arch}, the closed-world classifier training process integrates a 3D CNN and a RF.  A 3D CNN is used in this architecture in order to convolve over the space and the time dimensions. The use of 3D CNN for spatio-temporal data is already explored in~\cite{Ji2010,Tran2015}. 

The 3D CNN works as a feature extractor. In a first step, it is trained with the training data set $\mathcal{D}$. Then the fully-connected layer is removed and the flatten output before the fully-connected layer is denoted as $\boldsymbol{q}$. The CNN output $\boldsymbol{q}_j$ is then used as input for a RF. The output of the RF is $\textsl{\textbf{{\^{y}}}}$, an estimate of $\textsl{\textbf{{y}}}$. Based on the output of trees in a fully-grown RF, the $i$-th element of  $\textsl{\textbf{\^{y}}}$ is computed as the ratio of the number of trees that voted for class $c_i$ to the total number of trees $B$. The $i$-th element of $\textsl{\textbf{{\^{y}}}}_j$ for a datapoint $\boldsymbol{q}_j$ is

%
\begin{equation}\label{Eq:RFconf}
	\hat{P}(c_i|\boldsymbol{q}_j,\boldsymbol{\theta}_\mathrm{r})=\frac{1}{B}\sum_{b=1}^{B}\hat{P}_b(c_i|\boldsymbol{q}_j,\boldsymbol{\theta}_\mathrm{r}),
\end{equation}
where $\hat{P}_b(c_i|\boldsymbol{q}_j,\boldsymbol{\theta}_\mathrm{r}) \in \{0,1\}$ is the decision of $b$-th tree in RF in favor of class $c_i$ and $\boldsymbol{\theta}_\mathrm{r}$ is the set of RF parameters. 

\subsection{Vote-Based Unknown Class Detection Model}\label{SuSec:OST}
The aim of the open-set training is to learn the model for detecting unknown classes based on the 3D CNN, the RF trained as in Section \ref{sec:closeset} and the calibration dataset. As indicated in Eq.~(\ref{Eq:RFconf}), a data sample is classified as class $c_i$ if the number of trees voting for class $c_i$ is higher than that for other classes i.\,e., majority voting. The vote pattern of trees in the RF provides comprehensive information which is explored in this work to learn an unknown class detection model, named \textit{vote-based EVT Model}.    

\subsubsection{Vote Pattern and Extreme Value}
If a data sample does not belong to any of the known classes, then the number of trees voting for each of the known classes is expected to be small since the trees are trained for known classes only. 

To explain this phenomenon, the MNIST~\cite{LeCun2010} dataset of handwritten digits is used as a toy example. It has $10$ classes, each representing images of digits $0-9$. For the purpose of OSR in this work, the data samples in MNIST labelled with digits $0-5$ are treated as six $known$ classes and the digits $6-9$ as one $unknown$ class. For data samples of known classes in MNIST, they are split into a training dataset ($70\%$), a calibration dataset ($10\%$) and a testing dataset ($20\%$). The CNN\footnote{a simple VGG Net is used in this case} and the RF with $B=200$ trees are trained, as explained in Section \ref{sec:closeset}. 

\begin{figure}[h!]
	\centering
	\includegraphics[width=0.8\columnwidth]{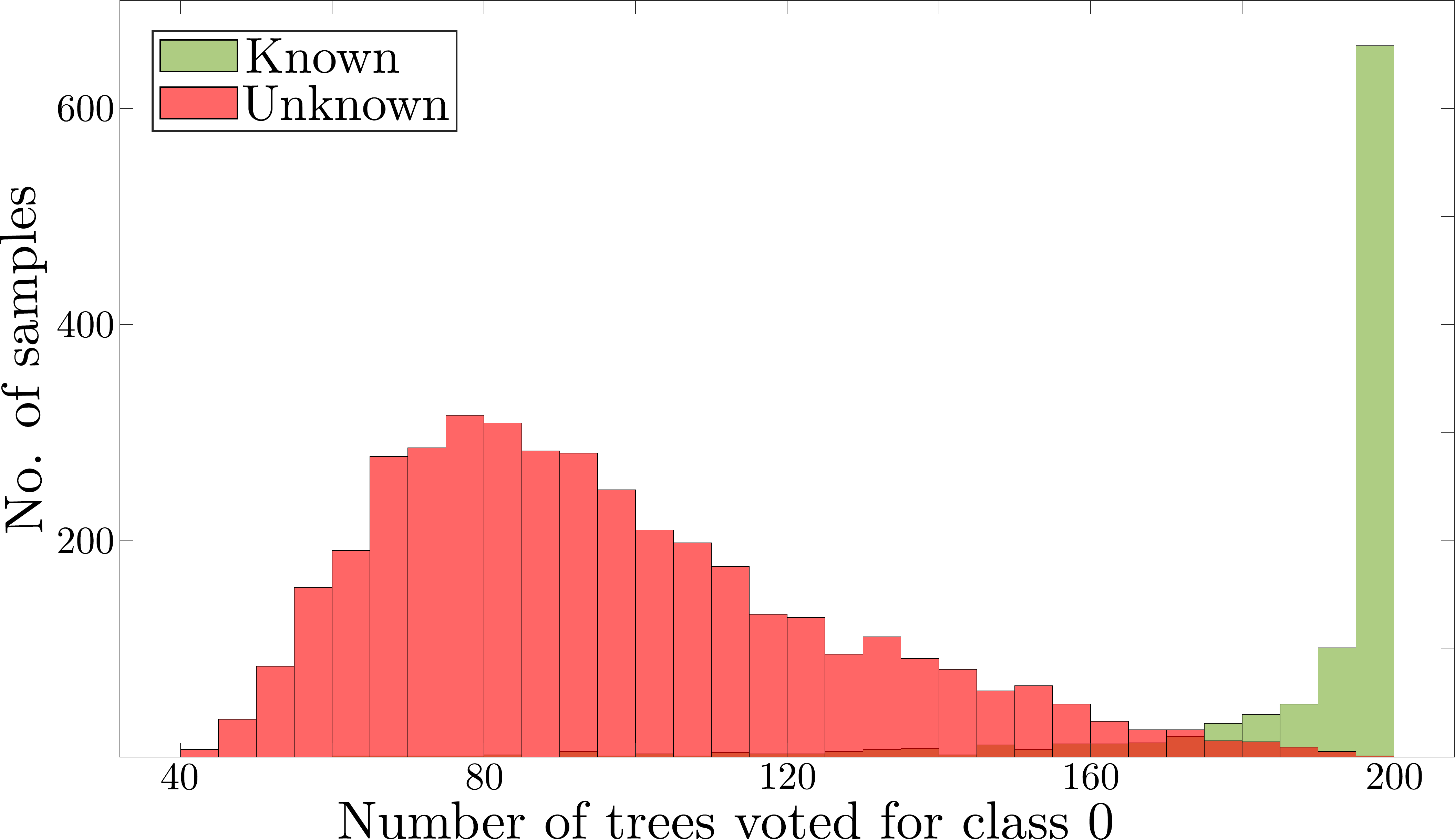}
	\caption{Vote-based unknown class detection.}
	\label{Fig:sd}	
\end{figure}

In the calibration dataset, there are $1000$ data samples labelled with the digit $0$. For each of such data samples, the number of trees in the RF voting for the digit $0$ is counted. The histogram for these data samples is illustrated in Fig.~\ref{Fig:sd} with the green bins. The $x$-axis represents the number of trees and $y$-axis the number of data samples. From Fig.~\ref{Fig:sd}, the green bins show that about $650$ data samples labelled as digit $0$ are correctly voted by $190-200$ trees, and $100$ data samples are correctly voted by $180-190$ trees, etc. Given a data sample from the unknown class, i.\,e., digits $6-9$, it is interesting for OSR to know how to reject such a data sample when it is classified by majority vote as digit $0$. 

The data points from the unknown class i.\,e., class 6-9 are also processed through the same pipeline. For each of them, the number of trees in the RF voting for the digit $0$ is determined. As shown in Fig.~\ref{Fig:sd} by the red bins, about $370$ data samples are voted as digit $0$ by about $80$ trees, about $150$ data samples are voted as digit $0$ by about $120$ trees, about $50$ data samples are voted as digit $0$ by about $160$ trees, etc. 

It can be seen in Fig.~\ref{Fig:sd}, that only a very small fraction of data samples belonging to the unknown class are voted as digit $0$ with over $180$ trees in the RF. Meanwhile, only a very small fraction of data samples belonging to the digit $0$ are voted as digit $0$ with less than $180$ trees. In other words, the number of votes for digit $0$ is a good indicator if a data sample belongs to the digit $0$ or not. If a data sample is rejected as not belonging to any of the known classes, i.\,e., the digits $0-5$, it likely belongs to the unknown class.
\begin{figure}[h!]
	\centering
	\begin{subfigure}{0.45\columnwidth}
		\includegraphics[width=\columnwidth]{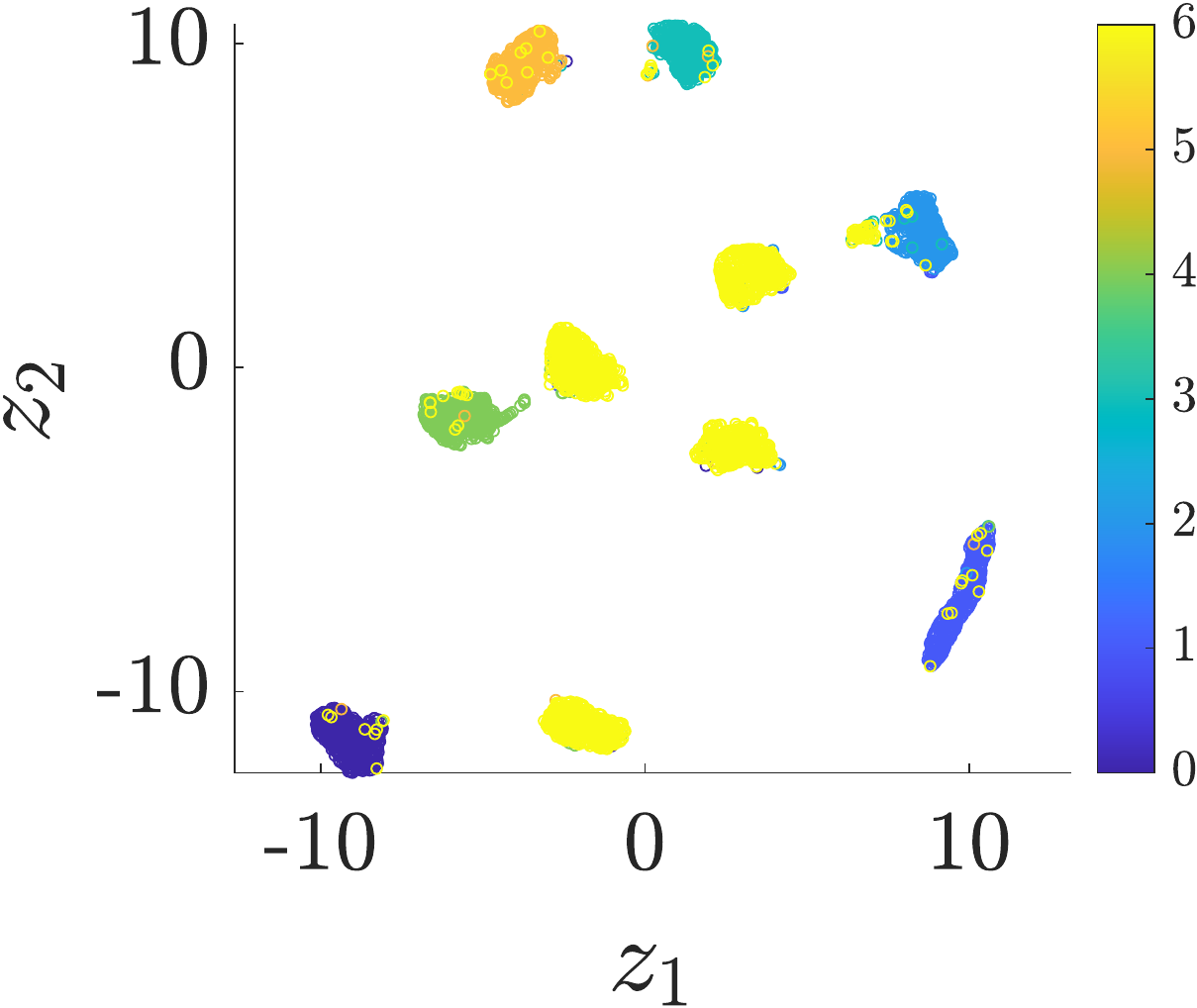}
		\caption{}
		\label{Fig:target_plot}
	\end{subfigure}
	\begin{subfigure}{0.45\columnwidth}
		\includegraphics[width=\columnwidth]{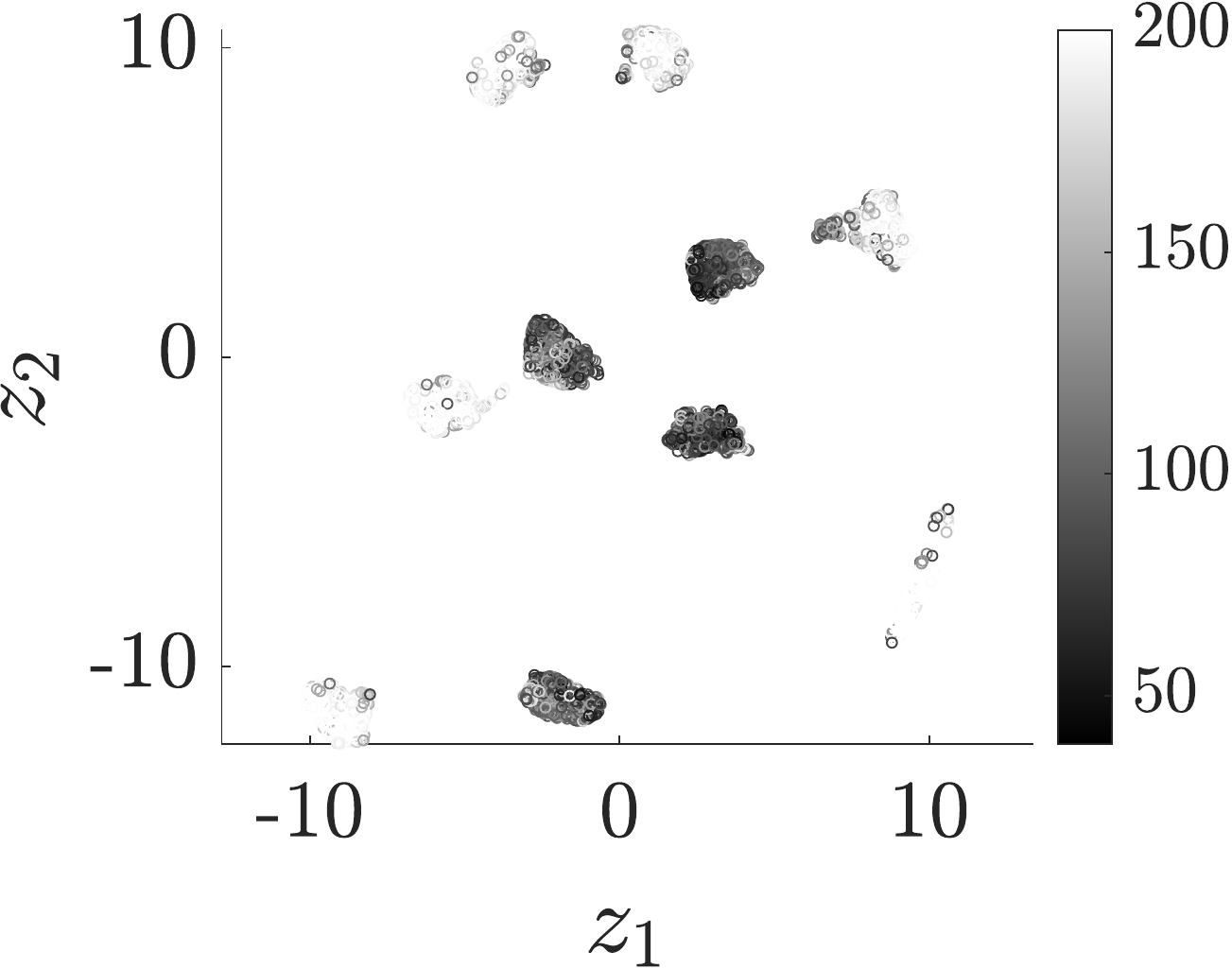}
		\caption{}
		\label{Fig:treeumaps}
	\end{subfigure}
	\caption{UMAP projection of known and unknown classes from the features extracted from MNIST trained only 0-5 classes.}
	\label{Fig:Mnistxempler1} 
\end{figure}
This can be further verified using the Uniform Manifold Approximation and Projection (UMAP),~\cite{McInnes2018UMAPUM}  by projecting the features extracted by the CNN for the MNIST samples from the calibration dataset to two dimensions as shown in Fig.~\ref{Fig:target_plot} and Fig.~\ref{Fig:treeumaps}. Similarly, the digits $0-5$ are treated as known classes, and the digits $6-9$ as an unknown class, i.\,e., the class $6$. Fig.~\ref{Fig:target_plot} shows the UMAP projection with the digits $0-5$ coloured according to the classes $0-5$. The digits $6-9$ are coloured yellow and marked as class $6$. Fig.~\ref{Fig:treeumaps} shows the same projection but coloured according to the number of trees in the RF voting for the corresponding class. Since the RF is trained on known classes, i.\,e., the digits $0-5$, the number of trees voting for $0-5$ is very large, in average close to the total number of trees, $B=200$. In contrast, the number of trees voting for the unknown class, i.\,e., class 6, is very small, in average close to $50$. These results implicate that the number of votes in a RF are well-suited to distinguish known classes from unknown classes.

During the CNN and RF training, there is no access to the data samples of unknown classes. However, the above observations disclose that data samples can be rejected as not belonging to a known class, if the number of trees in the RF voting for each of the known classes is small. This motivates the proposed \textit{vote-based EVT model}.

\subsubsection{Vote-Based EVT Model}\label{sec:evtmodel}
Extreme Value Theory is a branch of statistics dealing with the extreme deviations from the median of probability distributions. It seeks to assess, from a given ordered data sample of given random variables, the probability of events that are more extreme than any previously observed. Extreme value analysis is widely used in many disciplines, such as structural engineering, finance, earth sciences, traffic prediction, and geological engineering. In this work, EVT is used to estimate the probability that a data sample belongs to a class $c_k$ or an unknown class  based on the number of trees in the RF who voted for supporting the classification.  

For example, for the green bins in Fig.~\ref{Fig:sd}, the tail of the green bins can be defined as the left side of 180 intuitively. The core idea is that if a data sample does not belong to the digit $0$ it is very likely that less than $180$ trees vote for the class $0$. The tail boundary is defined $\lambda\cdot B$, where $B$ is the number of trees in RF and $\lambda \in (0,1)$ is a hyper-parameter. For example, if there are $200$ trees and $\lambda=0.9$, the tail boundary is $180$. For this purpose, extreme value distributions are used to model the values below the tail boundary and provide class specific statistics for rejection.

In this work, a two parameter Weibull distribution, which is a special case of the generalized extreme value distribution, is adopted. For each of the $K$ known classes, class specific Weibull distribution is modeled to capture the extreme values, i.\,e., the tail, in the vote patterns of trees in the RF as illustrated in Fig.~\ref{Fig:sd}. The $K$ established Weibull distributions together constitute the Vote-based EVT Model. For a known class $c_i$, the probability density function of a Weibull distribution for a  random variable $x$ is:
\begin{equation}\label{Eq:weibull}\centering
	\mathcal{W}_i(x;\alpha_i,\gamma_i) =
	\begin{cases}
		\frac{\alpha_i}{x}\left(\frac{x}{\alpha_i}\right)^{\gamma_i}e^{-(x/\alpha_i)^{\gamma_i}} & x\geq0 ,\\
		0 & x<0,
	\end{cases}
\end{equation}
where $\gamma_i>0$ is the shape parameter and $\alpha_i > 0$ is the scale parameter of the distribution. The parameters $\gamma_i$ and $\alpha_i$ are estimated using maximum likelihood estimation.

To generate class specific Weibull distributions $\mathcal{W}_1,\ldots,\mathcal{W}_K$ as shown in the Fig.~\ref{Fig:Arch} for $K$  training classes, the calibration dataset is used. The calibration dataset consists of data samples only from the $K$ known classes. For all the data points that are classified correctly to one of the $K$ known classes, the number of trees that voted for each correctly classified class is collected in the sets ${S}_{1},\ldots,{S}_K$, with $S_k=\{s_k^1,\ldots,s_k^{M_k}\}$, where,  and  $M_k$ is the number of scenarios belonging to class $c_k$ in the calibration dataset. For a class $c_k$, as only the extreme values are of interest, ${S}_k$ is sorted in the ascending order and only vales below the tail boundary from the sorted ${S}_k$ are chosen for Weibull modelling. To set the tail boundary, first a filter using  $\lambda$ is done to remove highly confident data points which cannot belong to the unknown class based on confidence score from Eq.~(\ref{Eq:RFconf}),  i.\,e., $\hat{P}\text{($c_k$}|\boldsymbol{q}_j,\boldsymbol{\theta}_\mathrm{r}\text{)}< \lambda$. The remaining data points in $S_k$ are used for estimating the class specific Weibull parameters $\alpha_k$ and ~$\gamma_k$ by maximum likelihood estimation.


\subsection{Unknown Class Prediction}\label{sec:prediction}
As discussed, the 3D CNN and RF have been trained using the training dataset, and the vote-based EVT model i.\,e., class specific Weibull models, has been learned using the calibration dataset. Recall that the training and calibration datasets contain data samples of known classes only. The test data, or any new data samples collected in an application, includes data samples of both known and unknown classes. For each data sample in the test data, the 3D CNN, RF and voted-based EVT model are used to predict whether it belongs to the unknown class or not.   

Given a data sample $\textbf{G}_j$ in the test dataset, it is first processed using the 3D CNN and the RF. Then, it is processed by the vote-based EVT model. As discussed, the vote-based EVT model consists of $\mathcal{W}_i(\alpha_i,\gamma_i)$ for all $K$ known classes. The cumulative probability that $\textbf{G}_j$ belongs to class $c_i$
is calculated by:

\begin{equation}
	\label{Eq:weibullCDF}
	\centering
	P(c_i|s_i^j,\alpha_i,\gamma_i)=
	\begin{cases}
		1-e^{-(s_i^j/\alpha_i)^{\gamma_i}},& \text{if}\  s_i^j > 0\\
		0,                     & \text{otherwise.}
	\end{cases}
\end{equation}
where $s_i^j$ is the number of trees in the RF voting for class $c_i$. Given a threshold $\delta$, if $P(c_i|s_i^j,\alpha_i,\gamma_i)<\delta$ then the sample is classified as not belonging to class $c_i$. If a data sample does not belong to any of the known $K$ classes, it is classified as unknown. In this work, experiments show that $\delta=0.5$ is a suitable value. 
\begin{figure}[h!]
	\centering
	\begin{subfigure}{0.45\columnwidth}
		\includegraphics[width=\columnwidth]{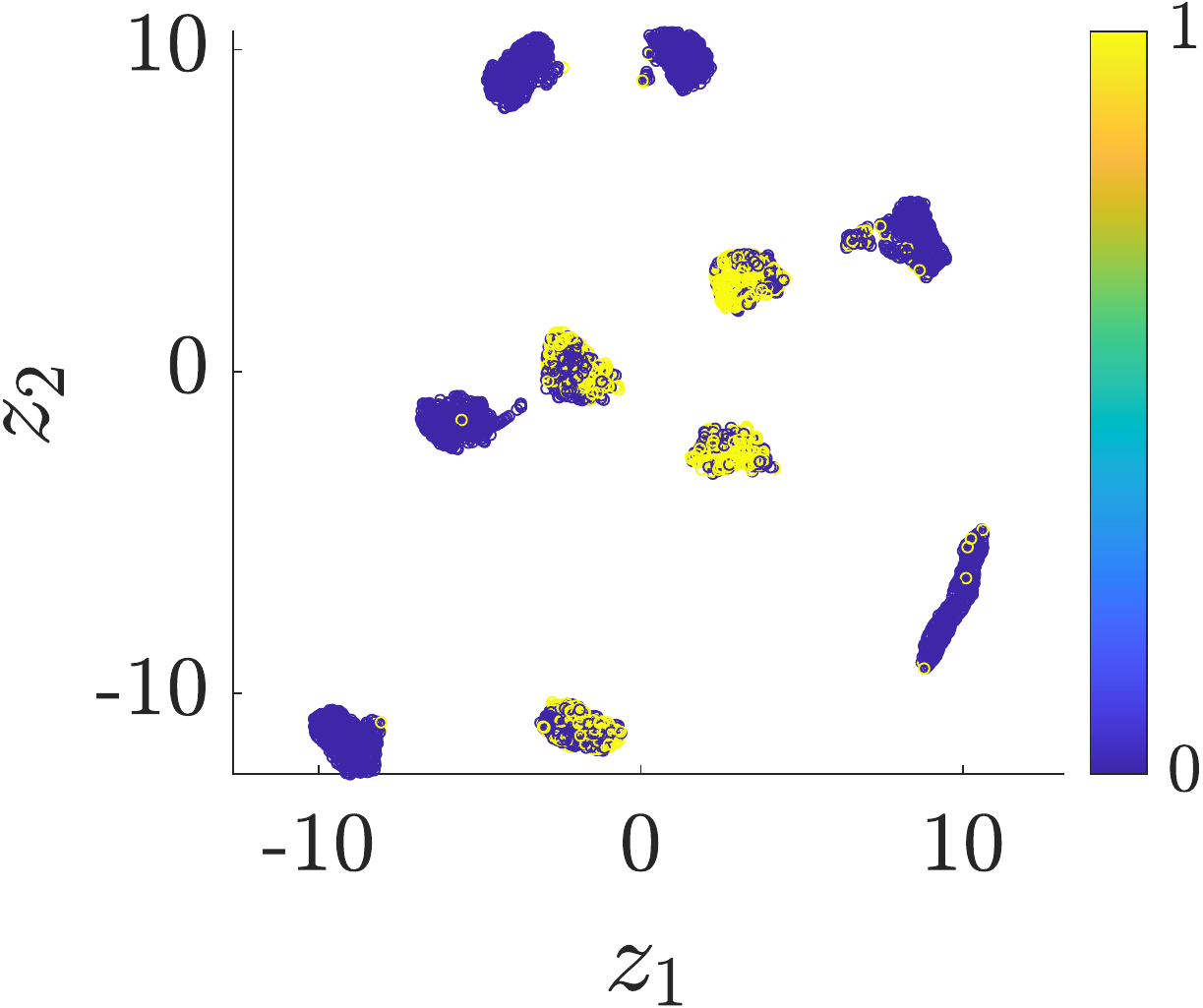}
		\caption{}
		\label{Fig:tree_binary_plot}
	\end{subfigure}
	\begin{subfigure}{0.45\columnwidth}
		\includegraphics[width=\columnwidth]{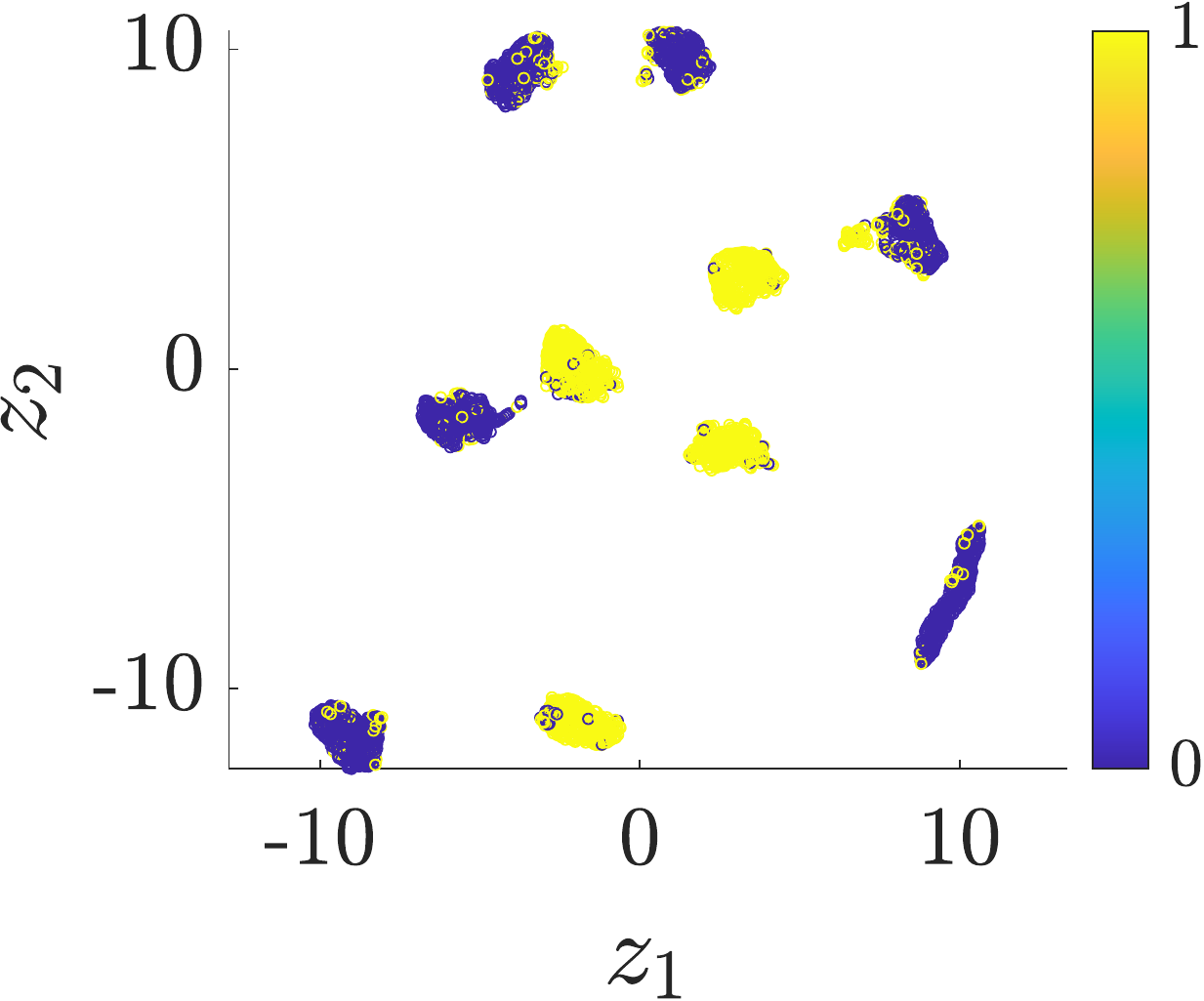}
		\caption{}
		\label{Fig:evt_binary_plot}
	\end{subfigure}
	\caption{CNN+RF trained on MNIST using classes $0-5$ and the classes $6-9$ are treated as unknown. Here blue represents known classes and yellow represents unknown class}
	\label{Fig:Mnistxempler} 
\end{figure}

To underline the advantage of the EVT based procedure, the MNIST OSR classification experiment trained with $6$ known and $4$ unknown classes as mentioned above is used. Fig.~\ref{Fig:target_plot} shows the ground truth projection with the known classes color coded according to the labels and the unknown classes color coded as yellow.  Fig.~\ref{Fig:tree_binary_plot} shows the result when a data sample is assigned to the unknown class (label 1) if the number of trees voting for the class with the highest value $\hat{P}(c_i|\boldsymbol{q}_j,\boldsymbol{\theta}_\mathrm{r})$ from Eq.~(\ref{Eq:RFconf}) is less than $50\%$. It can be observed that some data samples of unknown classes are mistakenly classified as some known class. Fig.~\ref{Fig:evt_binary_plot} shows the situation when Eq.~(\ref{Eq:weibullCDF}) is used instead of majority voting. The data samples are labelled as 1 (unknown class) if $P(c_i|s_i^j,\alpha_i,\gamma_i)<0.5$. It can be observed that the data samples of the unknown class that are mistakenly classified as some known class have been significantly reduced.  

\begin{figure}[h!]
	\centering
	\begin{subfigure}{0.07\columnwidth}
		\includegraphics[width=\columnwidth]{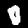}
	\end{subfigure}
	\begin{subfigure}{0.07\columnwidth}
		\includegraphics[width=\columnwidth]{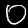}
	\end{subfigure}
	\begin{subfigure}{0.07\columnwidth}
		\includegraphics[width=\columnwidth]{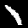}
	\end{subfigure}
	\begin{subfigure}{0.07\columnwidth}
		\includegraphics[width=\columnwidth]{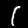}
	\end{subfigure}
	\begin{subfigure}{0.07\columnwidth}
		\includegraphics[width=\columnwidth]{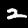}
	\end{subfigure}
	\begin{subfigure}{0.07\columnwidth}
		\includegraphics[width=\columnwidth]{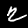}
	\end{subfigure}
	\begin{subfigure}{0.07\columnwidth}
		\includegraphics[width=\columnwidth]{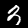}
	\end{subfigure}
	\begin{subfigure}{0.07\columnwidth}
		\includegraphics[width=\columnwidth]{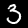}
	\end{subfigure}
	\begin{subfigure}{0.07\columnwidth}
		\includegraphics[width=\columnwidth]{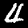}
	\end{subfigure}
	\begin{subfigure}{0.07\columnwidth}
		\includegraphics[width=\columnwidth]{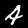}
	\end{subfigure}
	\begin{subfigure}{0.07\columnwidth}
		\includegraphics[width=\columnwidth]{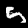}
	\end{subfigure}
	\begin{subfigure}{0.07\columnwidth}
		\includegraphics[width=\columnwidth]{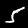}
	\end{subfigure}
	\caption{Images from the $known$ classes that are classified as $unknown$.}
	\label{Fig:Mnistxempler2} 
\end{figure}

With the vote-based EVT model, some data samples of known classes may also be classified as unknown. Fig.~\ref{Fig:Mnistxempler2} shows some of such data samples in MNIST where $P(c_i|s_i^j,\alpha_i,\gamma_i)<0.5$. These data samples are different from the usual way of handwritten digits and it might be hard to assign the right class. 

\section{Experiments}
\label{Sec:Exper}
This section presents the experiments conducted on real-world driving datasets and the results are also analysed. As in~\cite{Bendale2014,Shu2017,Yoshihashi2019,Oza2019a}, there are two ways to prepare datasets for testing OSR solutions. The first is \textit{class selection} which randomly selects a subset of classes among all classes in the dataset as the known classes, and the remaining classes together are considered as the unknown class. The second way is \textit{outlier addition} which takes all classes in one dataset as the known classes, and the data samples from another dataset area assigned as the unknown class. Both of these methods will be used for testing the proposed method.  

\subsection{Dataset}

\subsubsection{highD}

\begin{figure}[h!]
	\centering
	\includegraphics[width=0.5\linewidth]{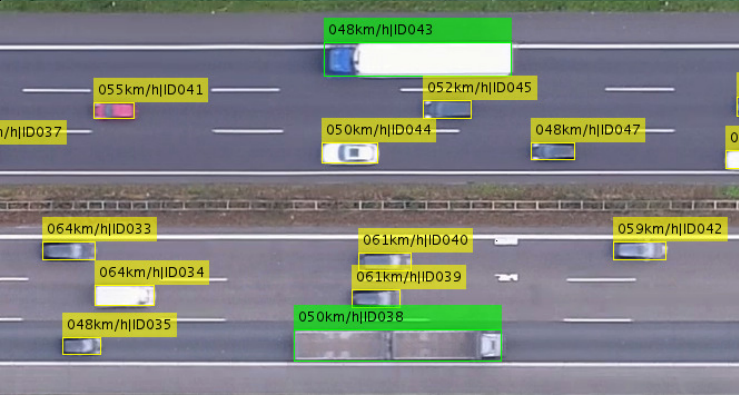}
	\caption{Illustration of the drone image with vehicles from the highD. The image is from the publication~\cite{Krajewski2018a}.}
	\label{Fig:highD}
\end{figure}

The highD dataset has a series of drone based video recordings of German multi-lane highway as shown in Fig.~\ref{Fig:highD}. Vehicle features and trajectories are also extracted and given as part of the dataset. The dataset has around $110500$ vehicle recordings and a total drive time of $447$ h. 

\begin{figure*}[h!]
\vspace{2mm} 
	\centering
	\begin{subfigure}{0.5\columnwidth}
		\includegraphics[width=\columnwidth]{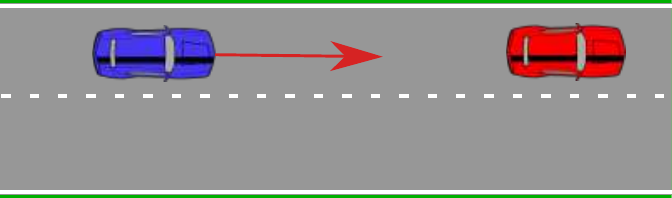}
		\caption{Ego - Following}
		\label{Fig:ego_follow}
	\end{subfigure}
	\begin{subfigure}{0.5\columnwidth}
		\includegraphics[width=\columnwidth]{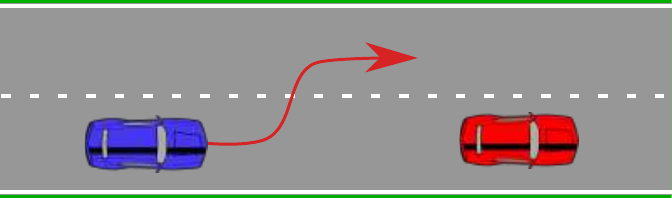}
		\caption{Ego - Left Lane Change}
		\label{Fig:ego_left}
	\end{subfigure}
	\begin{subfigure}{0.5\columnwidth}
		\includegraphics[width=\columnwidth]{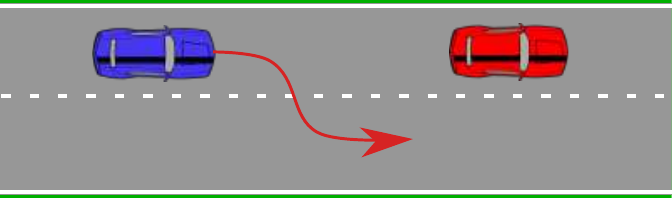}
		\caption{Ego - Right Lane Change}
		\label{Fig:ego_right}
	\end{subfigure}
	\begin{subfigure}{0.5\columnwidth}
		\includegraphics[width=\columnwidth]{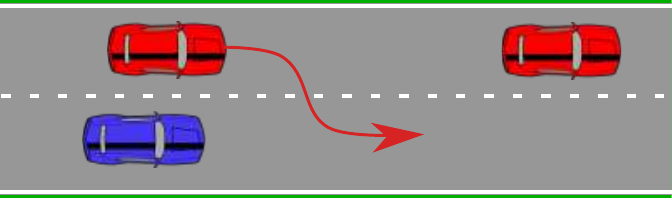}
		\caption{Leader - Cutin from Right}
		\label{Fig:cutin_right}
	\end{subfigure}
	\begin{subfigure}{0.5\columnwidth}
		\includegraphics[width=\columnwidth]{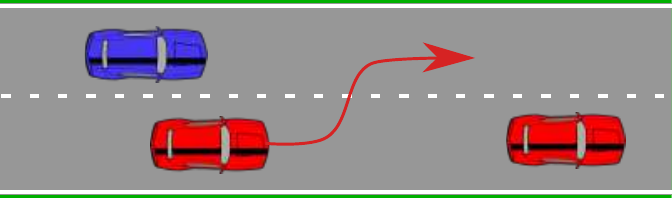}
		\caption{Leader - Cutin from Left}
		\label{Fig:cutin_left}
	\end{subfigure}
	\begin{subfigure}{0.5\columnwidth}
		\includegraphics[width=\columnwidth]{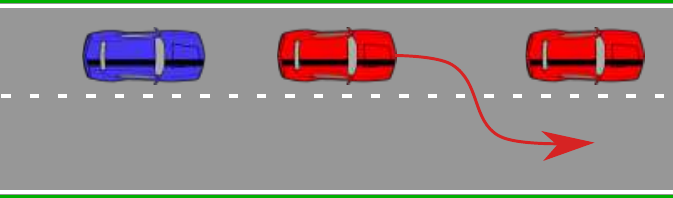}
		\caption{Leader - Cutout from Right}
		\label{Fig:cutout_right}
	\end{subfigure}
	\begin{subfigure}{0.5\columnwidth}
		\includegraphics[width=\columnwidth]{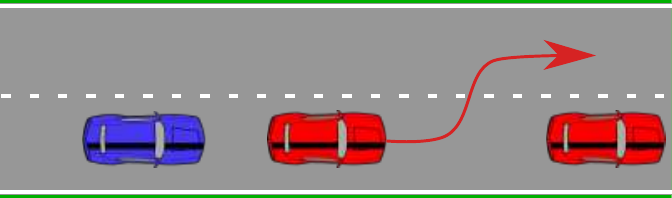}
		\caption{Leader - Cutout from Left}
		\label{Fig:cutout_left}
	\end{subfigure}
	\caption{Traffic scenario classes from highD. Here, the blue car is the ego vehicle and red cars are the traffic participants.}
	\label{Fig:highDClasses} 
\end{figure*}
As the focus of this work is on traffic scenarios classification, for validating the presented method $7$ common highway scenario classes as shown in Fig.~\ref{Fig:highDClasses} are extracted from the highD dataset. The $7$ classes are as follows:

\begin{itemize}
	\item Ego - Following
	\item Ego - Right lane change
	\item Ego - Left lane change
	\item Leader - Cutin from left 
	\item Leader - Cutin from right
	\item Leader - Cutout to left
	\item Leader - Cutout to right
\end{itemize}

As described in~\ref{subsec:traffic}, THW$ < 4$\unit{s} is used as the  criterion for finding relevant scenarios where ego vehicle has a leading vehicle. The environment at each time instance is represented with $\Occupancy$ of span $15$\unit{m}~$\times~200$\unit{m} and a resolution of $0.5$\unit{m}~$\times~1$\unit{m}. The interval  $t_0-t_\text{lb}$ and $\Delta t$ used in this work are 1.8\unit{s} and 0.2\unit{s} respectively. So, a traffic scenario is represented with \textbf{G} of size $30\times200\times10$. The grids are generated in an ego-centric fashion. In total 4480 scenarios were extracted, where the split of dataset is $70\%$  for training, $10\%$ for calibration and $20\%$ for testing.  

\subsubsection{OpenTraffic}

\begin{figure}[h!]
	\centering
	\includegraphics[width=0.5\linewidth]{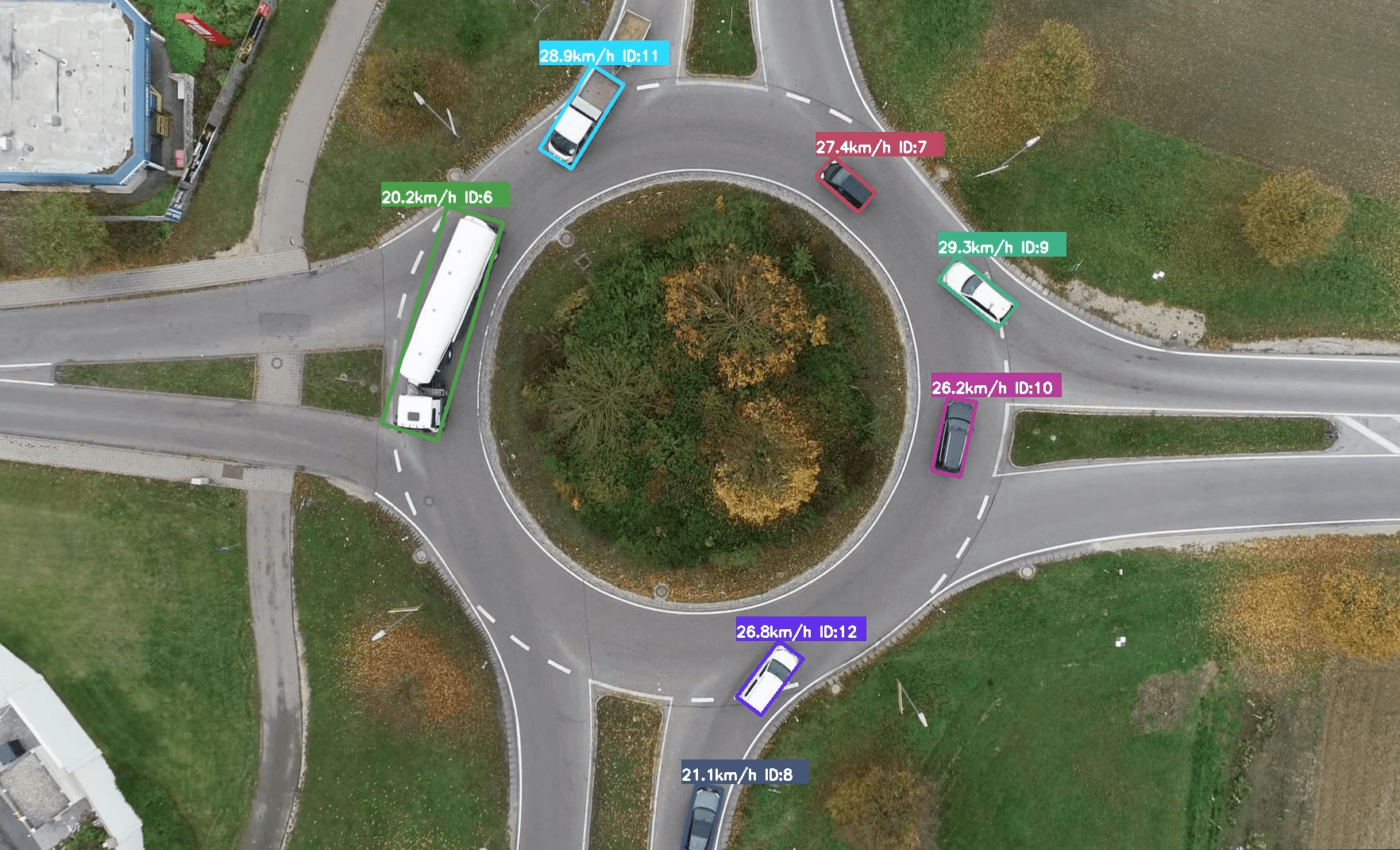}
	\caption{Illustration of the drone image with vehicles from the OpenTraffic. The image is from the publication~\cite{Kruber2020}.}
	\label{Fig:round}
\end{figure}

OpenTraffic is also a drone based dataset. Unlike highD, OpenTraffic focuses on roundabouts as shown in Fig~\ref{Fig:round}. Similar to highD, as a post processing step vehicle features and trajectories can be extracted. The OpenTraffic dataset is not labelled and will be used for the \textit{outlier addition} experiment, where all the data samples from the OpenTraffic dataset are considered as one unknown class. 
The scenario extraction and representation of the OpenTraffic dataset is also done in the same way as the highD dataset as mentioned above.

\subsection{Implementation Details}
The 3D CNN architecture used in this work is similar to the one used in~\cite{Ji2010} for video classification. The architecture details are given in the Table.~\ref{Tab:CNN_Architecthure}. For training, Adam~\cite{Kingma2014} optimiser is used and a batch size of 32 is chosen. The RF for the vote-based method is trained with $B=200$ trees. But as shown in the ablation study in Section~\ref{Sec:InfB}, $B$ has no influence on OSR performance atleast from the values $100$ and above. If the output of the CNN has $L$ features, i.\,e., $\boldsymbol{q}\in \mathbb{R}^L$, the RF is fully-grown by selecting $\sqrt{L}$ features as suggested in~\cite{Breiman2001} to find the split rule for each node in each tree of the RF. In the vote-based EVT model, the hyper-parameter $\lambda$ is $0.9$ as discussed in Section \ref{sec:evtmodel} and $\delta$ is set to be $0.5$ as discussed in Section \ref{sec:prediction}. The influence of the parameter $\delta$ is also discussed in the ablation study.    
\begin{table}[h!]

\centering
	\begin{tabular}{ |c|c|c|c| } 
		\hline
		layer & kernel & stride & input size \\
		\hline
		conv1 & $4\times12\times3$ & $1\times1\times1$ & $30\times200\times10\times1$ \\ 
		maxpool1 & $3\times3\times1$ & $1\times1\times1$ & $27\times189\times8\times8$ \\ 
		dropout1 ($0.25$) & - & - & $9\times63\times8\times8$ \\ 
		conv2 & $4\times8\times3$ & $1\times1\times1$ &  $9\times63\times8\times8$  \\ 
		maxpool2 & $2\times3\times1$ & $1\times1\times1$ &  $6\times56\times6\times6$  \\ 
		dropout2 ($0.25$) & - & - &  $3\times18\times6\times6$\\ 
		conv3 & $4\times12\times3$ & $1\times1\times1$ &  $3\times18\times6\times6$ \\
		flatten & - & - & 1020 \\
		dense & - & - & 500 \\
		dense & - & - &  $K$ \\
		\hline
	\end{tabular}
\caption{3D CNN architecture details.}
		\label{Tab:CNN_Architecthure}
\end{table}

\subsection{Baselines} The baseline methods used in this work to compare the OSR performance  are  
\begin{itemize}
    \item Openmax~\cite{Bendale2015}, a re-calibrated softmax score based on the per class EVT distribution of the class-wise feature distances
    \item DOC~\cite{Shu2017}, a normal CNN based classification network with a sigmoid layer at the end for OSR
    \item Softmax (Naive) and RF-conf (Naive), a rejection threshold of $0.5$ is applied on the maximum softmax score and confidence score from Eq.~(\ref{Eq:RFconf}) to distinguish unknown class and known classes
\end{itemize}   
Compared to works mentioned above, the presented work in this paper is an ensemble based and the spread of vote pattern among the ensembles is used as a measure of uncertainty. 

\subsection{Evaluation Metrics}
The Macro-average of $F$-score is used as the metric for evaluating the OSR solutions. The $F$-score is the harmonic mean of precision and recall,
\begin{equation}
	F\text{-score} = 2 \cdot \frac{\text{precision} \cdot \text{recall}}{\text{precision} + \text{recall}}
\end{equation}

where, recall $= \frac{TP}{TP+FN}$ and recall  $= \frac{TP}{TP+FP}$. Here, $TP,FP$ and $FN$ are the true positives, false positives and false negatives respectively. The Macro-average of $F$-score is computed by taking the average of the $F$-scores for each class. The score values vary from $0$ to $1$, with $1$ being the best score. 

\subsection{Class Selection}\label{sec:cs}
This section reports the experiments where the highD dataset is prepared by the \textit{class selection} method. Among the $7$ labelled classes, $4$ of them are retained as known and the other $3$ classes are treated as the \textit{unknown} class. The class selection is repeated $5$ times and each time the $4$ known classes are randomly selected. The 3D CNN, RF and the vote-based models are trained only on the $4$ known classes. The ratio of known to unknown classes are also varied and the results are analysed in the ablation study. The Macro-average of $F$-score with the standard deviations are presented in Table~\ref{Tab:highDClassSelec}. The proposed vote-based EVT Model is compared against baselines: Openmax, DOC, Softmax (Naive) and RF-conf (Naive). A larger score and a smaller standard deviation indicate better performance. The results show that the proposed vote-based EVT Model provides better and reliable performance in the conducted experiment. 

\begin{table}[h]
	\centering
	\caption{Macro-average of $F$-score (class selection).}
	\begin{tabular}{| c | c |} 
		\hline
		Method &  highD - 4 Known vs 3 Unknown  \\ 
		\hline\hline
		Softmax (Naive) & 0.642$\pm$0.013 \\  
		Openmax & 0.71$\pm$0.017 \\
		DOC & 0.692$\pm$0.019  \\
		Rf-conf (Naive) &  0.635$\pm$0.012 \\
		Vote-based EVT & \textbf{0.77$\pm$0.017} \\	
		\hline
	\end{tabular}
	\label{Tab:highDClassSelec}
\end{table}

\subsection{Outlier Addition}
This section reports the experiments where the datasets are prepared by the \textit{outlier addition} method. In this experiment, all the labelled classes from the highD dataset are treated as known classes, and combined with the so-called noise dataset in which the samples from a different dataset are treated as one unknown class. The noise-dataset used in this work is from the OpenTraffic dataset. The number of test samples from the highD dataset is matched with the number of test samples from the OpenTraffic dataset. Hence, the number of known to unkown test samples is $1:1$. The results of this experiment are presented in the Table~\ref{Tab:highDOltlit}. The proposed vote-based EVT model outperforms all baselines significantly on the outlier dataset.

\begin{table}[h]
	\centering
	\caption{Macro-average of $F$-score (outlier addition).}
	\begin{tabular}{| c | c |} 
		\hline
		Method &  OpenTraffic  \\ 
		\hline\hline
		Softmax (Naive) & 0.733 \\  
		Openmax & 0.92 \\
		DOC & 0.91 \\
		Rf-conf (Naive) & 0.9044 \\
		Vote-based EVT & \textbf{0.931} \\	
		\hline
	\end{tabular}
	\label{Tab:highDOltlit}
\end{table}
\section{Ablation Study}
\label{Sec:Abla}
\subsubsection{Ratio of Known to Unknown Classes}
\begin{figure}[h!]
	\centering
	\includegraphics[width=0.9\columnwidth]{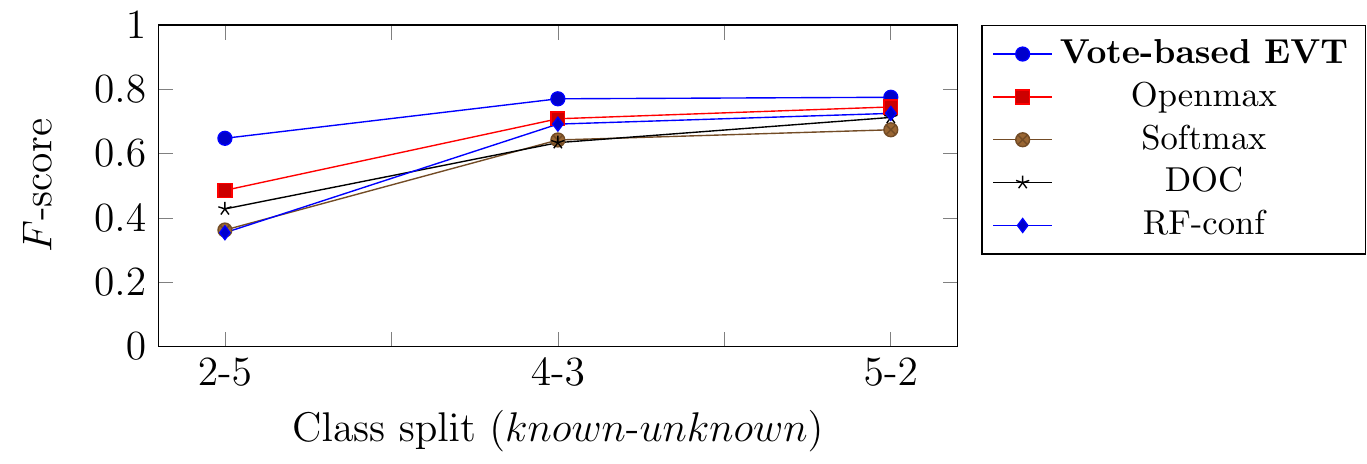}
	\caption{Class split vs $F-$score}
	\label{Fig:classsplit}
\end{figure}

So far, all test with the class selection method are performed on datasets where there are $7$ labelled classes. Among them, $4$ classes are used as known classes and the remaining $3$ classes are treated as one unknown class. So, the ratio of known to unknown classes is $4:3$. To evaluate the robustness of the OSR methods, the analysis concerning this ratio is briefly reported. The tests are performed at different ratios of known to unknown classes: $2:5,~   5:2$. Clearly, the performance of all methods degrades at low ratio, i.\,e., $2:5$ and increases when the ratio increases. The results are shown in Fig.~\ref{Fig:classsplit}. For all ratios, the proposed vote-based EVT model has demonstrated robust performance in spite of the problem becoming more open.
\subsubsection{Influence of $\delta$}
\begin{figure}[h!]
	\centering
	\includegraphics[width=0.9\columnwidth]{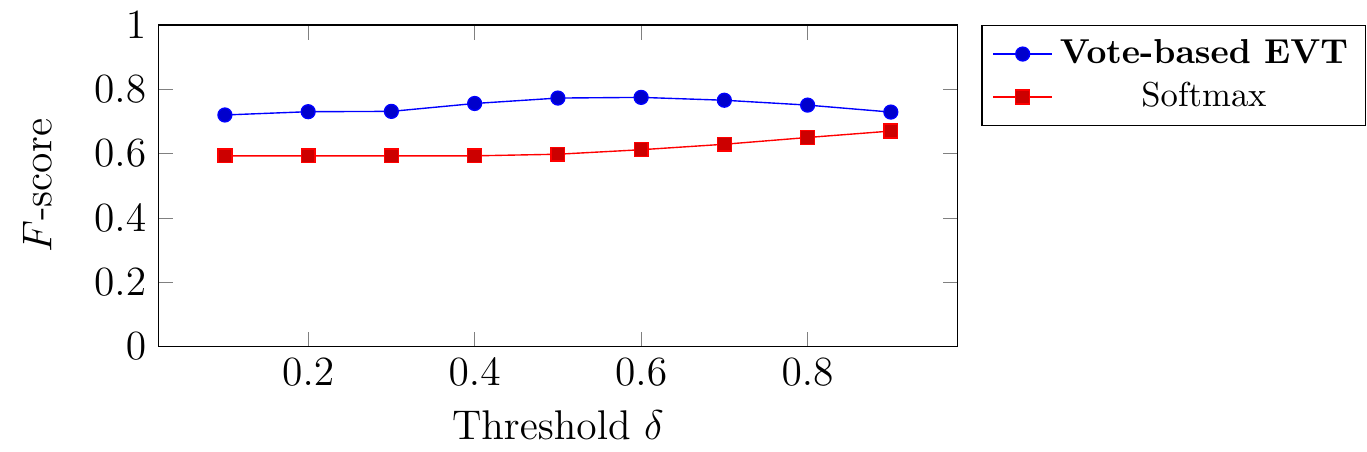}
	\caption{Threshold $\delta$ vs $F-$score}
	\label{Fig:ThresholdGraph}
\end{figure}

The influence of the threshold parameter $\delta$ of the vote-based EVT model for the OSR performance  is analysed by varying $\delta$ from $0.1$ to $1$. For comparison, the Softmax score is also varied from $0.1$ to $0.9$ and the OSR performance is reported. For the above mentioned study, the known to unknown class ratio is kept at $4:3$ on the highD dataset and all other parameters are kept the same as mentioned in the Sec.~\ref{Sec:Exper}. The results are shown in Fig.~\ref{Fig:ThresholdGraph}. The results show that the proposed vote-based EVT score is better than the softmax score for classifying traffic scenarios to one of the known classes or as an unknown class. Also, the OSR performance is robust for several values of $\delta$.
\begin{figure}[h!]
	\centering
	\includegraphics[width=0.9\columnwidth]{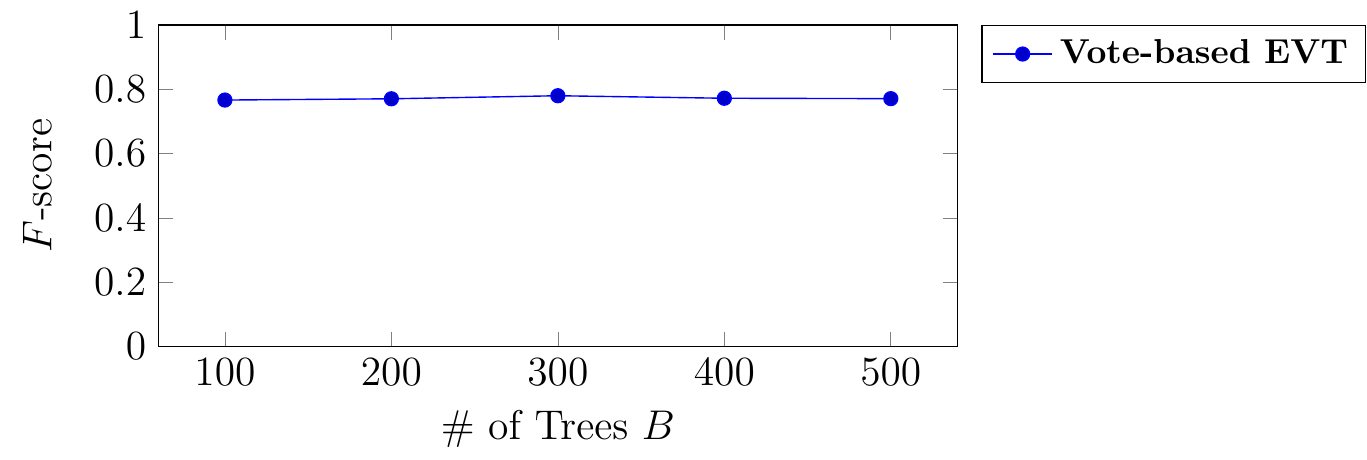}
	\caption{\# of Trees $B$ vs $F-$score}
	\label{Fig:TreeGraph}
\end{figure}
\subsubsection{Influence of $B$}
\label{Sec:InfB}
The clout of the number of trees $B$ in the RF for the OSR performance  is analyzed by varying it from $100$ to $500$. The known to unknown class ratio is kept at $4:3$ on the highD dataset for this study and all other parameters are kept the same as mentioned in the Sec.~\ref{Sec:Exper}.  

The results are shown in Fig.~\ref{Fig:TreeGraph}. From the figure below, it can be seen that the \# of trees has almost no influence on the OSR performance.

\section{Conclusions}
\label{sec:conclusion}
OSR is an open problem of traffic scenario classification even though it is of high practical importance for autonomous driving. This work proposes a combination of 3D CNN and the RF algorithm along with extreme value theory to address the OSR task. The proposed solution, named vote-based EVT method, is featured by exploring the vote patterns of decision trees in RF for known classes and applying EVT to detect the unknown classes. Extensive tests on highD and OpenTraffic datasets  have verified the advantages of the new solution compared to state-of-the-art methods.



.

\section*{Acknowledgment}

This work is supported by Bavarian State Ministry for Science and Art under the funding code VIII.2-F1116.IN/18/2.


\bibliographystyle{IEEEtran}
\bibliography{reference_new}

\end{document}